\documentclass[letterpaper, 10 pt, conference]{ieeeconf}  

\IEEEoverridecommandlockouts                              

\usepackage[english]{babel}
\usepackage{amsmath}
\usepackage{amssymb}
\usepackage{booktabs}
\usepackage[dvips]{graphicx}
\usepackage{multirow}
\usepackage{amsfonts}
\usepackage{enumerate}
\usepackage{tabularx}
\usepackage{algorithm,algorithmic}
\usepackage{bm}
\usepackage{enumerate}
\usepackage{adjustbox}
\usepackage{xcolor}
\usepackage{siunitx}
\usepackage{booktabs}
\usepackage{mdframed}
\usepackage{adjustbox}
\usepackage{authblk}
\usepackage{color}
\usepackage{xurl}
\usepackage{subcaption}
\usepackage{cite}
\makeatletter
\let\NAT@parse\undefined
\makeatother
\usepackage{hyperref}
\usepackage{relsize}
\usepackage{float}
\usepackage{pifont}

\newcommand{\xmark}{\ding{55}}%
\usepackage{wrapfig,lipsum}

\title{\LARGE \bf  UASTHN: Uncertainty-Aware Deep Homography Estimation for UAV Satellite-Thermal Geo-localization}

\author{Jiuhong Xiao and Giuseppe Loianno
\thanks{The authors are with New York University, Tandon School of Engineering, Brooklyn, NY 11201, USA. {\tt\footnotesize email: \{jx1190, loiannog\}@nyu.edu}.}
\thanks{This work was supported by the Technology Innovation Institute, the NSF CAREER Award 2145277, the NSF CPS Grant CNS-2121391, and the NYU IT High Performance Computing resources, services, and staff expertise. Giuseppe Loianno serves as consultant for the Technology Innovation Institute. This arrangement has been reviewed and approved by the New York University in accordance with its policy on objectivity in research.}
}

\begin{document}
\maketitle
\thispagestyle{empty}
\pagestyle{empty}

\begin{abstract}
Geo-localization is an essential component of Unmanned Aerial Vehicle (UAV) navigation systems to ensure precise absolute self-localization in outdoor environments. To address the challenges of GPS signal interruptions or low illumination, Thermal Geo-localization (TG) employs aerial thermal imagery to align with reference satellite maps to accurately determine the UAV's location. However, existing TG methods lack uncertainty measurement in their outputs, compromising system robustness in the presence of textureless or corrupted thermal images, self-similar or outdated satellite maps, geometric noises, or thermal images exceeding satellite maps. To overcome these limitations, this paper presents \textit{UASTHN}, a novel approach for Uncertainty Estimation (UE) in Deep Homography Estimation (DHE) tasks for TG applications. Specifically, we introduce a novel Crop-based Test-Time Augmentation (CropTTA) strategy, which leverages the homography consensus of cropped image views to effectively measure data uncertainty. This approach is complemented by Deep Ensembles (DE) employed for model uncertainty, offering comparable performance with improved efficiency and seamless integration with any DHE model. Extensive experiments across multiple DHE models demonstrate the effectiveness and efficiency of CropTTA in TG applications. Analysis of detected failure cases underscores the improved reliability of CropTTA under challenging conditions. Finally, we demonstrate the capability of combining CropTTA and DE for a comprehensive assessment of both data and model uncertainty. Our research provides profound insights into the broader intersection of localization and uncertainty estimation. The code and models are publicly available.
\end{abstract}

\section*{Supplementary Material}
{ \noindent \textbf{Project page:}  \url{https://xjh19971.github.io/UASTHN/}}
\section{Introduction}~\label{sec:introduction}
Over the past decade, Unmanned Aerial Vehicles (UAVs) have proven highly adaptable and efficient across various tasks, including agriculture~\cite{info10110349}, solar farm inspections~\cite{drones6110347}, search and rescue~\cite{atif2021uav}, power line monitoring~\cite{rao2022quadformer, powerline}, and object tracking~\cite{saviolo2023unifying}. Research has focused on improving UAV localization and navigation to ensure stable flight and accurate trajectory tracking. For long-term outdoor operations, accurate GPS localization is essential to prevent drift~\cite{review_avl}. When GPS is unreliable due to signal loss or interference, visual geo-localization~\cite{foundloc, vgscience, directalign3, imgregistration} aligns aerial images with satellite maps for robust positioning. In low-light conditions, Thermal Geo-localization (TG), using aerial thermal imagery~\cite{lee2024caltech}, image retrieval~\cite{stl}, and deep homography estimation~\cite{STHN}, supports effective navigation.

Despite the promising results of TG in aligning aerial thermal images with satellite imagery using deep learning techniques~\cite{lecun2015deep}, several challenges hinder their practical applications. First, these methods lack mechanisms to indicate low confidence when confronted with textureless or self-similar patterns in thermal or satellite images. Second, the reliance on north alignment through IMU and compass data, with limited tolerance for geometric noise, renders these systems susceptible to large geometric distortions. Third, existing approaches assume the availability of corresponding satellite images for global matching and that the UAV is within the search area, leading to failures if the UAV moves beyond this zone. Hence, incorporating uncertainty measurement is vital for improving inference reliability. 

\begin{figure*}[]
    \centering
    \includegraphics[width=\textwidth]{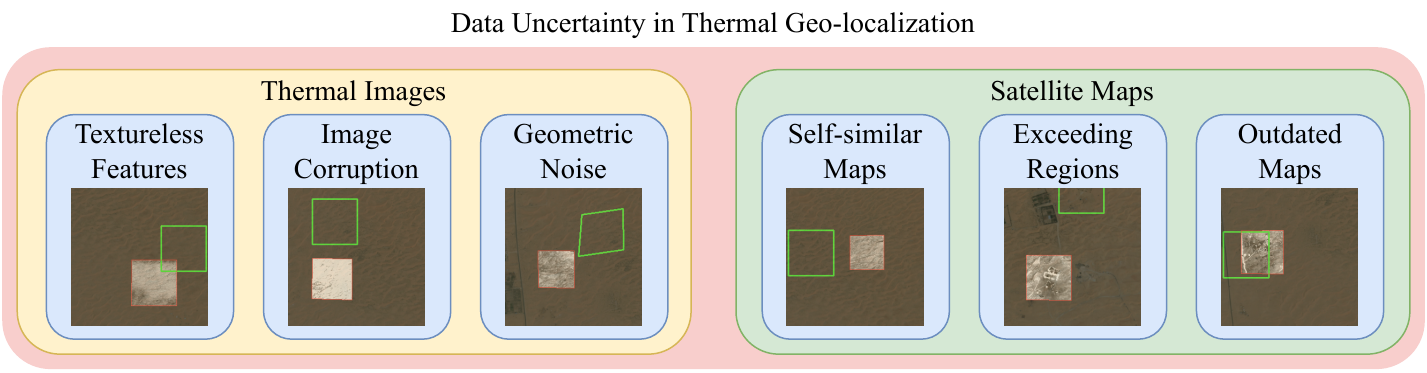}
    \caption{Data Uncertainty in Thermal Geo-localization (TG): Our approach captures six categories of high data-uncertainty samples leading to TG failure, where \textcolor{red}{predicted} displacements significantly deviate from the \textcolor{green}{ground truth}. Thermal images are overlaid on predicted displacements on the satellite imagery. High-resolution images are available on our project page.}
    \label{teaser}
    \vspace{-15pt}
\end{figure*}

Fig.~\ref{teaser} highlights key categories of high data uncertainty samples commonly encountered in TG, identified by our method: (a) \textbf{Textureless Features}: Low-contrast, textureless thermal images, especially at nighttime. (b) \textbf{Image Corruption}: Overexposed, underexposed, or noisy thermal data. (c) \textbf{Geometric Noise}: Severe north-alignment errors from inaccurate IMU or compass information causing geometric distortion. (d) \textbf{Self-similar Maps}: Repetitive satellite patterns (e.g., desert dunes) leading to false matches. (e) \textbf{Exceeding Regions}: Thermal images extending beyond satellite map. (f) \textbf{Outdated Maps}: Satellite images not reflecting recent developments, causing inconsistencies with thermal imagery.

In this study, we introduce \textit{Uncertainty-Aware Satellite-Thermal Homography Network (UASTHN)}, a sample and consensus-based Uncertainty Estimation (UE) framework for Deep Homography Estimation (DHE) in satellite-thermal geo-localization.Our main contributions are as follows:
\begin{itemize}
\item We introduce a CropTTA strategy with a unique homography consensus mechanism for effective data uncertainty measurement, seamlessly integrating with any DHE approach. Our study also comprehensively assesses model uncertainty using Deep Ensembles (DE).
\item We show how the proposed method is the first solution to address the challenge of uncertainty estimation for localization using cross-domain data, achieving superior homography estimation and geo-localization performance compared to baselines.
\item Extensive experiments validate our approach's effectiveness and efficiency on challenging satellite-thermal datasets. Specifically, our method achieves a geo-localization error of $7~\si{m}$ with a $97\%$ success rate for uncertainty estimation within a $512~\si{m}$ search radius.
\end{itemize}

\section{Related Works}~\label{sec:relatedworks}
\textbf{UAV Thermal Geo-localization (TG).} TG involves using thermal cameras on UAVs in conjunction with satellite maps to extract the locations of thermal images and determine the UAV's position. Existing approaches employ two primary strategies: global matching~\cite{foundloc, stl} and local matching. This work focuses specifically on local matching methods.


Local matching methods use Deep Homography Estimation (DHE)~\cite{detone2016deep, cao2022iterative, nguyen2018unsupervised, shao2021localtrans} or keypoint matching\cite{pmlr-v155-achermann21a} to align thermal and satellite images. Some works~\cite{electronics12040788, electronics12214441} employ conditional GANs for visual-thermal homography, while \cite{STHN} uses two-stage iterative DHE for TG, offering near real-time performance but struggling with challenging alignment such as textureless thermal images. Our method integrates an uncertainty estimation mechanism with CropTTA, improving DHE resilience by flagging low-confidence cases, and enhancing TG system reliability and situational awareness.

\textbf{Uncertainty Estimation for Deep Learning.}
Uncertainty Estimation (UE)\cite{gawlikowski2023survey, kendall2017uncertainties}, also known as uncertainty quantification\cite{ABDAR2021243}, is vital in safety-critical deep-learning applications like UAV localization and navigation. Without accurate UE, a neural network cannot indicate the reliability of its outputs and may show overconfidence in cases of high data uncertainty (aleatoric), such as noisy data or sensor failures, or high model uncertainty (epistemic), like out-of-distribution samples, potentially leading to system failures. We examine the following categories of UE methods:

\textit{Test-Time Augmentation (TTA).} TTA methods~\cite{shanmugam2021better, kimura2021understanding, kim2020learning, zhang2022memo} use data augmentation during evaluation to combine outputs from augmented input samples and measure uncertainty. These methods explore the best augmentation tailored for different tasks and primarily target classification tasks but have limited application to regression tasks like DHE.

\textit{Deep Ensembles (DE).} DE methods~\cite{lakshminarayanan2017simple, NEURIPS2021_a70dc404,fort2019deep,abe2022deep} train multiple models with varying initializations and data orders, then combine their outputs to assess uncertainty. Although greater model diversity often enhances performance, its impact on out-of-distribution samples remains debated~\cite{fort2019deep, abe2022deep}.

For UE in deep homography estimation, existing works mainly use visibility masks~\cite{zhang2022hvc} and pixel-level photometric matching uncertainty~\cite{xu2022cuahn}. In contrast, our CropTTA method leverages the homography consensus of crop-augmented images to measure data uncertainty. This approach can seamlessly integrate with any DHE method and is proven to be both effective and efficient for TG.
\vspace{-5pt}


\section{Methodology}~\label{sec:methodology}
\begin{figure*}
    \centering
    \includegraphics[width=0.85\linewidth]{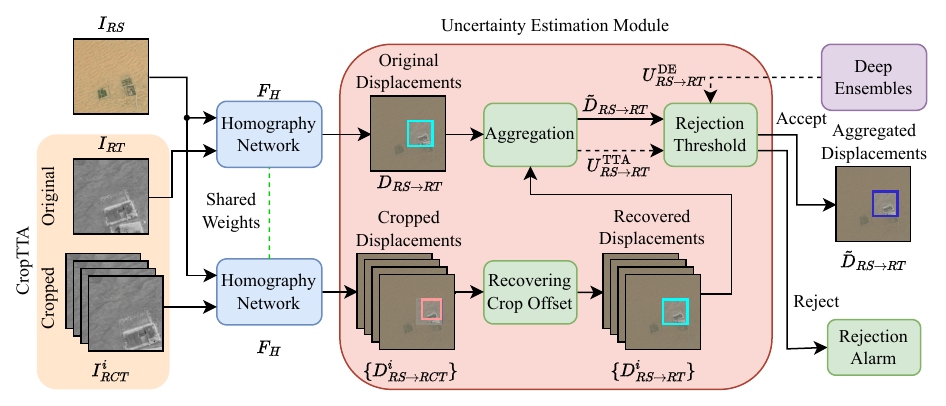}
    \caption{UASTHN framework: CropTTA augments thermal images, and network $F_H$ with an UE module calculates aggregated displacements ($\tilde D_{RS\rightarrow RT}$) and data uncertainty ($U^\textrm{TTA}_{RS\rightarrow RT}$). $U^\textrm{TTA}_{RS\rightarrow RT}$ is used to reject samples with high uncertainty. Optionally, DE estimates model uncertainty ($U^\textrm{DE}_{RS\rightarrow RT}$), which can be combined with CropTTA for comprehensive UE.}
    \label{fig:framework}
    \vspace{-15pt}
\end{figure*}

Our UASTHN framework is illustrated in Fig.~\ref{fig:framework}. The framework consists of two main components: a Deep Homography Estimation (DHE) module and an Uncertainty Estimation (UE) module utilizing Crop-based Test-Time Augmentation (CropTTA) and Deep Ensembles (DE).

\subsection{Deep Homography Estimation (DHE) Module}
The DHE module employs a homography network $F_H$. We denote $W_S$ as the size of the square satellite image $I_S$ and $W_T$ as the size of the square thermal image $I_T$. Both images are resized to $W_R$, yielding $I_{RS}$ and $I_{RT}$. The homography network $F_H$ takes $I_{RS}$ and $I_{RT}$ and outputs the four-corner displacement $D_{RS\rightarrow RT} \in \mathbb{R}^{2 \times 4}$, indicating the displacement from the four corners of $I_{RS}$ to $I_{RT}$. This displacement essentially aligns $I_{RT}$ into $I_{RS}$. Subsequently, Direct Linear Transformation (DLT)~\cite{ABDELAZIZ2015103} utilizes this displacement to compute the homography matrix $H_{RS\rightarrow RT}$.

\subsection{Crop-based Test-Time Augmentation (CropTTA)}

We propose CropTTA as a simple and effective method for measuring data uncertainty in DHE for TG. Our approach involves augmenting $I_T$ by cropping it with a specific crop offset $o_c$. We denote the $i^\text{th}$ augmented thermal image as $I^i_{CT}$ with the size of $W_{CT} = W_T - o_c$, where $i=1,\cdots,N_C-1$ and $N_C-1$ is the number of augmented samples for each $I_T$. We explore two sampling methods: random sampling and grid sampling. In random sampling, the samples consist of the original image and $N_C-1$ randomly cropped images. In contrast, the grid sampling method utilizes the original image and $N_C-1$ cropped images covering four corners.

Next, we resize $I^i_{CT}$ to $I^i_{RCT}$ to $W_R$, and composite the thermal image batch as $\{I_{RT}, I^1_{RCT},\cdots,I^{N_C-1}_{RCT}\}$. The homography network $F_H$ predicts displacements $\{D_{RS\rightarrow RT}, D^1_{RS\rightarrow RCT},\cdots,D^{N_C-1}_{RS\rightarrow RCT}\}$. To recover the displacements $D^i_{RS\rightarrow RT}$ from the cropped displacements $D^i_{RS\rightarrow RCT}$, we apply the following transformations
\begin{equation}
    H^i_{RS\rightarrow RT} = \textrm{DLT}(\mathbf{\hat x}^i_{RCT}, \mathbf{x}^i_{RCT}),
\end{equation}
\begin{equation}
    H^i_{RCT\rightarrow RT} = \textrm{DLT}(\mathbf{\hat x}^i_{RCT}, \mathbf{\hat x}_{RT}),
\end{equation}
\begin{equation}\small
     \begin{bmatrix}
       \bm{x}^i_{RT}\\ \bm{y}^i_{RT} \\ \mathbf{1}
   \end{bmatrix} = H^i_{RS\rightarrow RT}H^i_{RCT\rightarrow RT}(H^i_{RS\rightarrow RT})^{-1}\begin{bmatrix}
       \bm{x}^i_{RCT}\\ \bm{y}^i_{RCT} \\ \mathbf{1}
   \end{bmatrix},
\end{equation}
\begin{equation}\small
   \mathbf{x}^i_{RT} = \begin{bmatrix}
       \bm{x}^i_{RT}\\ \bm{y}^i_{RT}
   \end{bmatrix},
    \mathbf{x}^i_{RCT} = \begin{bmatrix}
       \bm{x}^i_{RCT}\\ \bm{y}^i_{RCT}
   \end{bmatrix} = \mathbf{x}_{RS} + D^i_{RS\rightarrow RCT},
\end{equation}
and we get recovered displacements as
\begin{equation}
        D^i_{RS\rightarrow RT} = \mathbf{x}^i_{RT} - \mathbf{x}_{RS},
\end{equation}
where $\mathbf{x}_{RS},~\mathbf{x}^i_{RT},~\mathbf{x}^i_{RCT}\in \mathbb{R}^{2\times4}$ represent the four-corner coordinates of $I_{RS}$, and the $i^\text{th}$ predicted four-corner coordinates of $I_{RT}$ and $I^i_{RCT}$ respectively. $\mathbf{\hat x}_{RT},~\mathbf{\hat x}^i_{RCT}\in \mathbb{R}^{2\times4}$ denote the four-corner coordinates of $I_{RT}$ and $I^i_{RCT}$ before the predicted homography transformation. Conversely,  $\mathbf{x}^i_{RCT}$ and $\mathbf{x}^i_{RT}$ are the coordinates after transformation. $H^i_{RCT\rightarrow RT}$ and $H^i_{RS\rightarrow RT}$ denote homography matrices from $I^i_{RCT}$ to $I_{RT}$ and from $I_{RS}$ to $I_{RT}$ for the $i^\text{th}$ prediction. $\bm{x}^i_{RT},~\bm{y}^i_{RT},~\bm{x}^i_{RCT},~\bm{y}^i_{RCT} \in \mathbb{R}^{1\times 4}$ are the $x$ and $y$ coordinates of $\mathbf{x}^i_{RT}$ and $\mathbf{x}^i_{RCT}$. We use the standard deviation (std) of displacements as the measurement of data uncertainty

\begin{equation}
    U^\textrm{TTA}_{RS\rightarrow RT} = \textrm{std}(\mathcal{D}_{RS\rightarrow RT}),
\end{equation}
\begin{equation}
\mathcal{D}_{RS\rightarrow RT} = \{D_{RS\rightarrow RT}, D^1_{RS\rightarrow RT},\cdots,D^{N_C-1}_{RS\rightarrow RT}\},
\end{equation} 
where $U^\textrm{TTA}_{RS\rightarrow RT}\in \mathbb{R}^{2\times4}$ represents the standard deviation for the estimated four-corner displacement. We denote the rejection threshold as $s_c$. The estimation results are rejected if $U^\textrm{TTA}_{RS\rightarrow RT} > s_c$, meaning all elements of $U^\textrm{TTA}_{RS\rightarrow RT}$ are larger than $s_c$. The aggregated displacement $\tilde D_{RS\rightarrow RT}$ is calculated using either the average displacements of all samples in $\mathcal{D}_{RS\rightarrow RT}$ or only the original displacement $D_{RS\rightarrow RT}$.

The intuition of CropTTA is that all cropped views share the same homography matrix as the original thermal image. If $I^\prime_T$ is generated from $I_T$ using matrix $H$, then for any cropped views $I_{CT}$, pixel coordinates transform as $x^\prime_{CT} = Hx_{CT}$, allowing calculation of the transformed four-corner coordinates. Additionally, when thermal images are affected by sensor issues or low-contrast inputs, the network $F_H$ predicts similar $D^i_{RS\rightarrow RCT}$. In such cases, $U^i_{RS\rightarrow RT}$ is dominated by $H^i_{RCT\rightarrow RT}$, maintained by sample distribution.

For the model training of the displacement with CropTTA, the loss function $\mathcal{L}_\textrm{CropTTA}$ is
\begin{equation}
\begin{split}
    \mathcal{L}_\textrm{CropTTA} &= \sum_{k=0}^{K_1-1}\gamma^{K_1 -k-1} \left( \Vert D_{k,RS \rightarrow RT} - D^{gt}_{k,RS \rightarrow RT}\Vert_1 \right.\\ & \left. + \sum_{i=1}^{N_C}\Vert D^i_{k,RS \rightarrow RT} -  D^{gt}_{k,RS \rightarrow RT}\Vert_1 \right),
\end{split}
\end{equation}
where $\gamma$ and $K_1$ denote the decay factor and the number of iterations for $F_H$ if $F_H$ is an iterative network; otherwise, $K_1 = 1$ and $\gamma=1.0$. $D^{gt}_{k,RS \rightarrow RT}$ is the ground truth.

\begin{figure*}[]
\begin{subfigure}[b]{0.33\textwidth}
    \includegraphics[width=\textwidth]{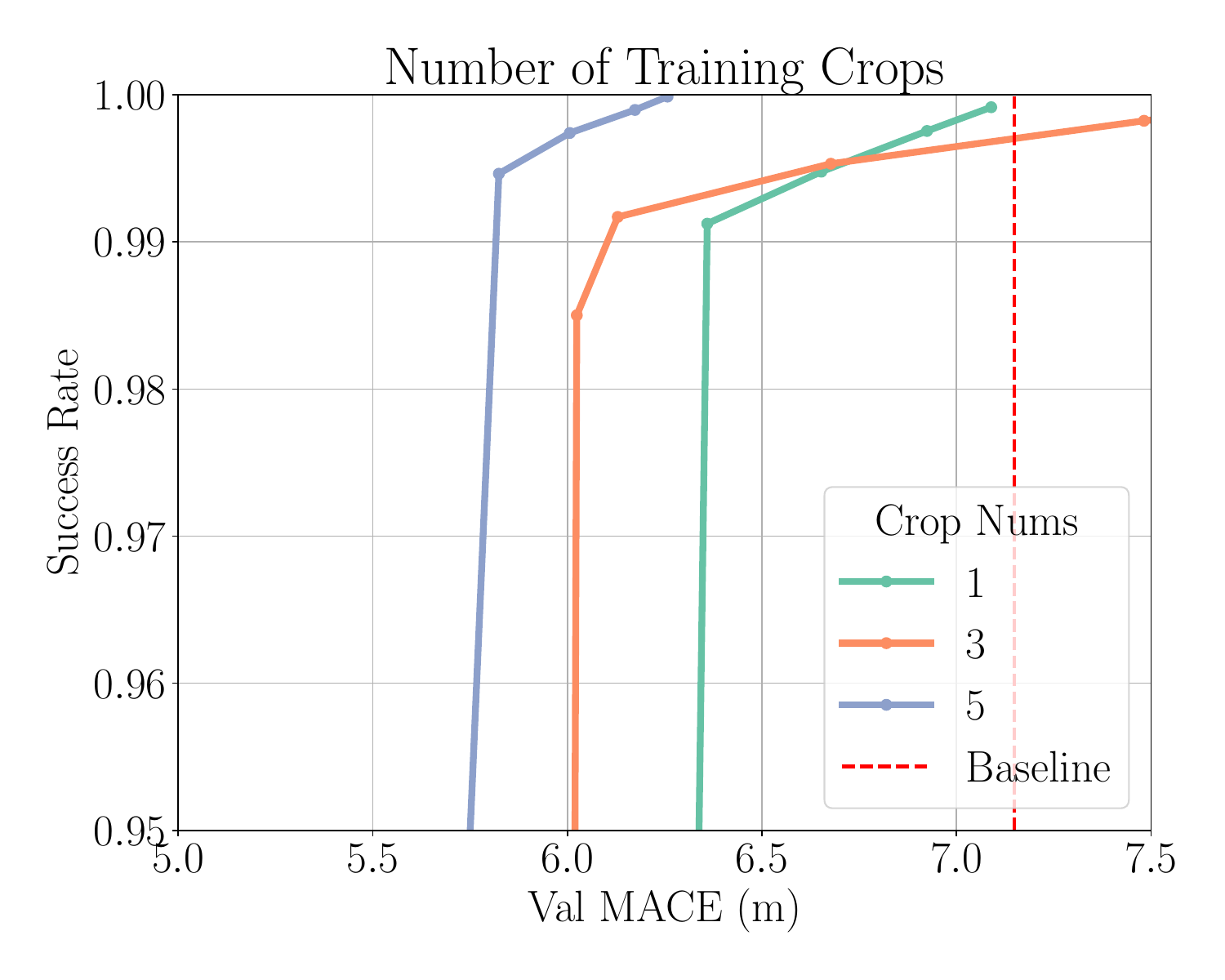}\vspace{-10pt}
    \caption{Number of Training Crops}
    
    \label{mct}
\end{subfigure}
\begin{subfigure}[b]{0.33\textwidth}
    \includegraphics[width=\textwidth]{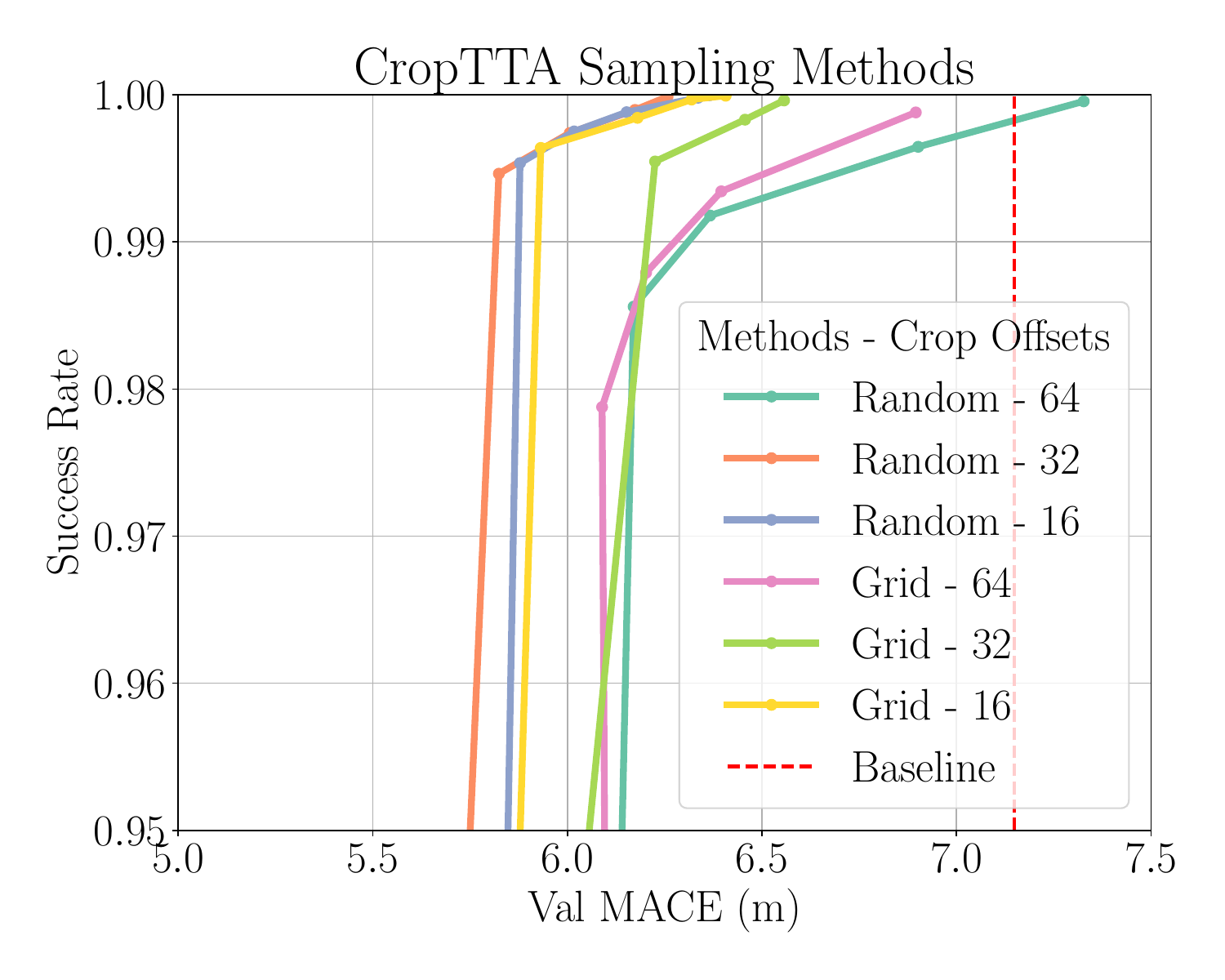}\vspace{-10pt}
    \caption{Sampling Methods}
    \label{croptta}
\end{subfigure}
\begin{subfigure}[b]{0.33\textwidth}
    \includegraphics[width=\textwidth]{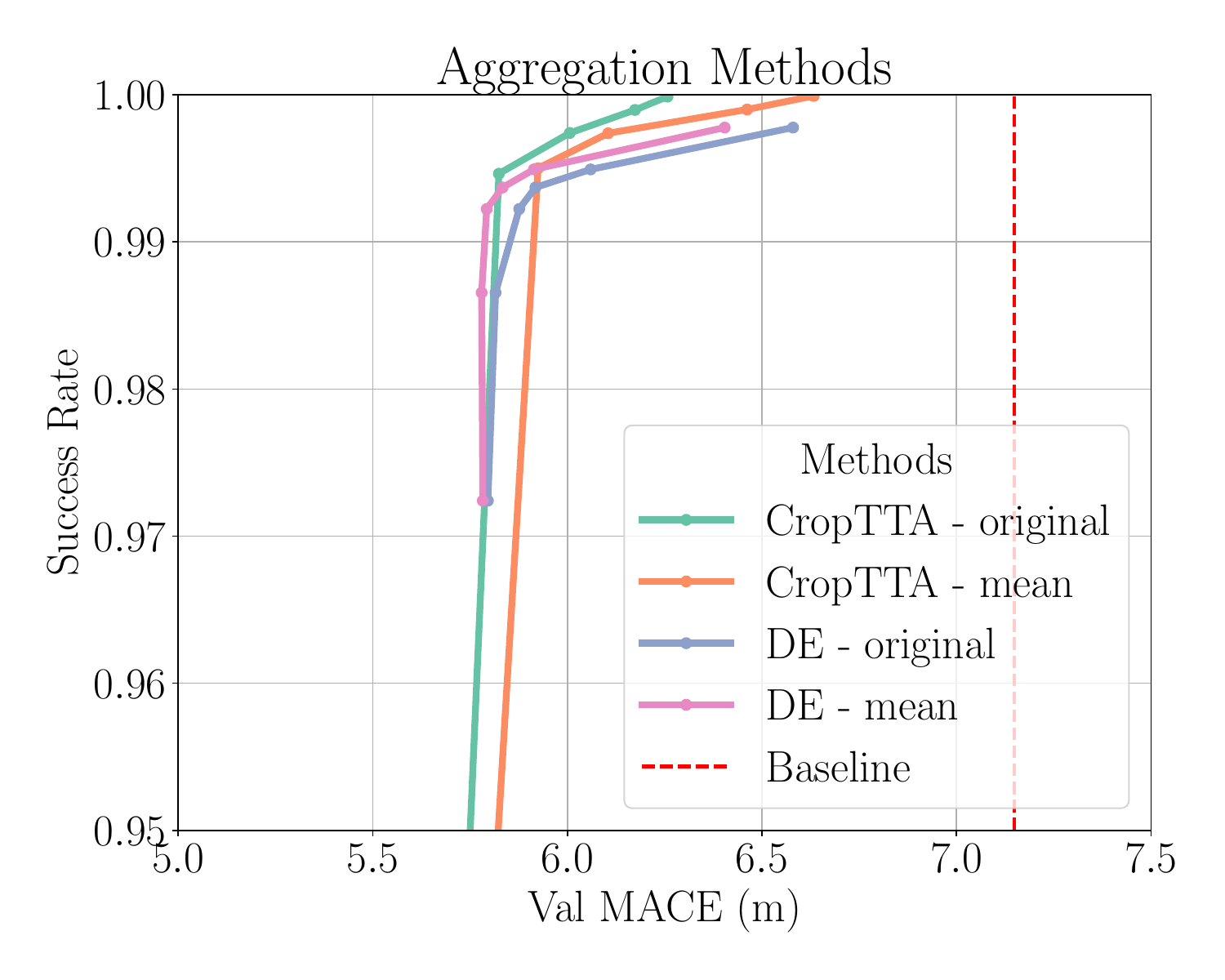}\vspace{-10pt}
    \caption{Aggregation Methods}
    \label{agg}
\end{subfigure}
\begin{subfigure}[b]{0.33\textwidth}
    \includegraphics[width=\textwidth]{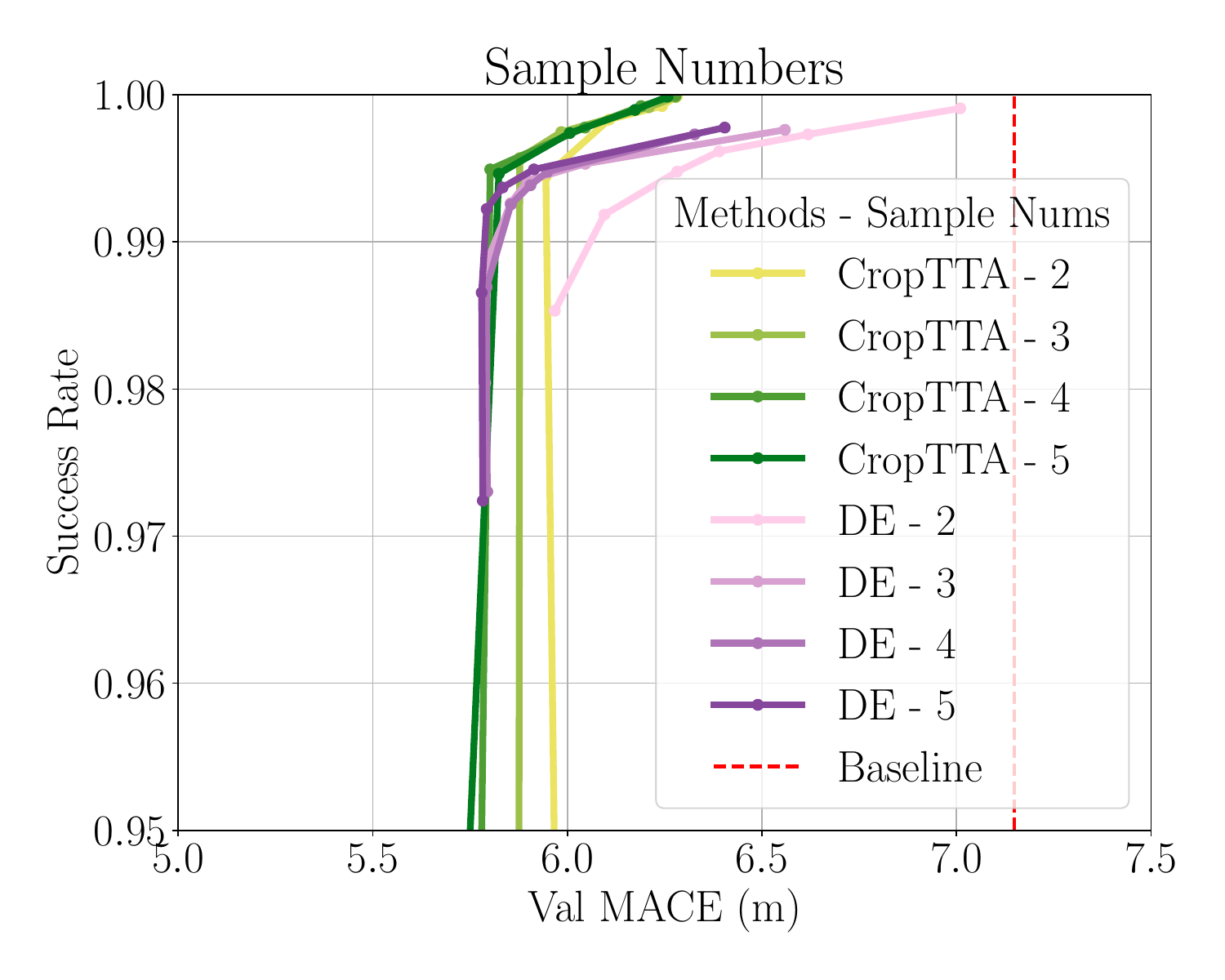}\vspace{-10pt}
    \caption{Sample Numbers}
    \label{sample}
\end{subfigure}
\begin{subfigure}[b]{0.33\textwidth}
    \includegraphics[width=\textwidth]{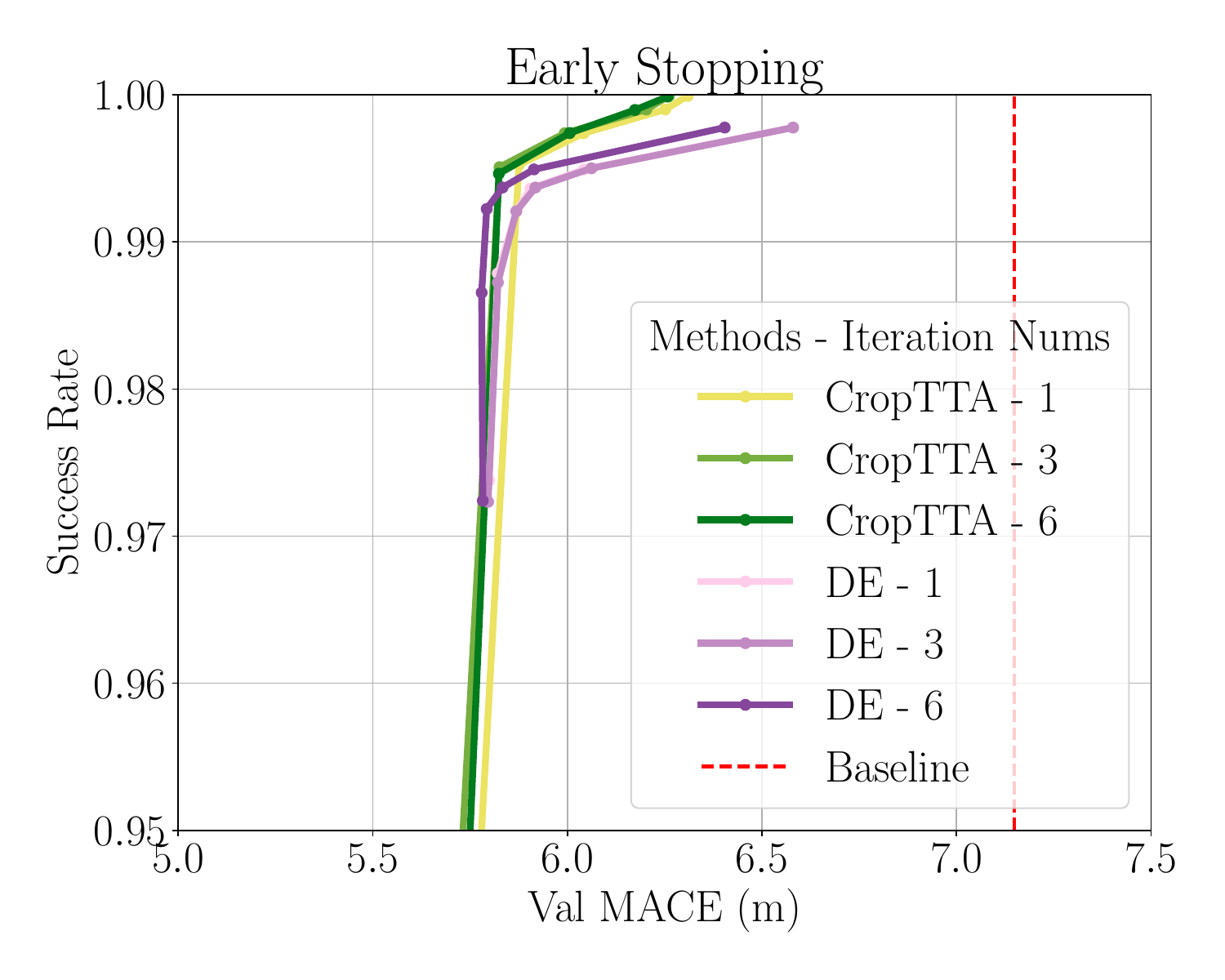}\vspace{-10pt}
    \caption{Early Stopping}
    \label{cs}
\end{subfigure}
\begin{subfigure}[b]{0.33\textwidth}
    \includegraphics[width=\textwidth]{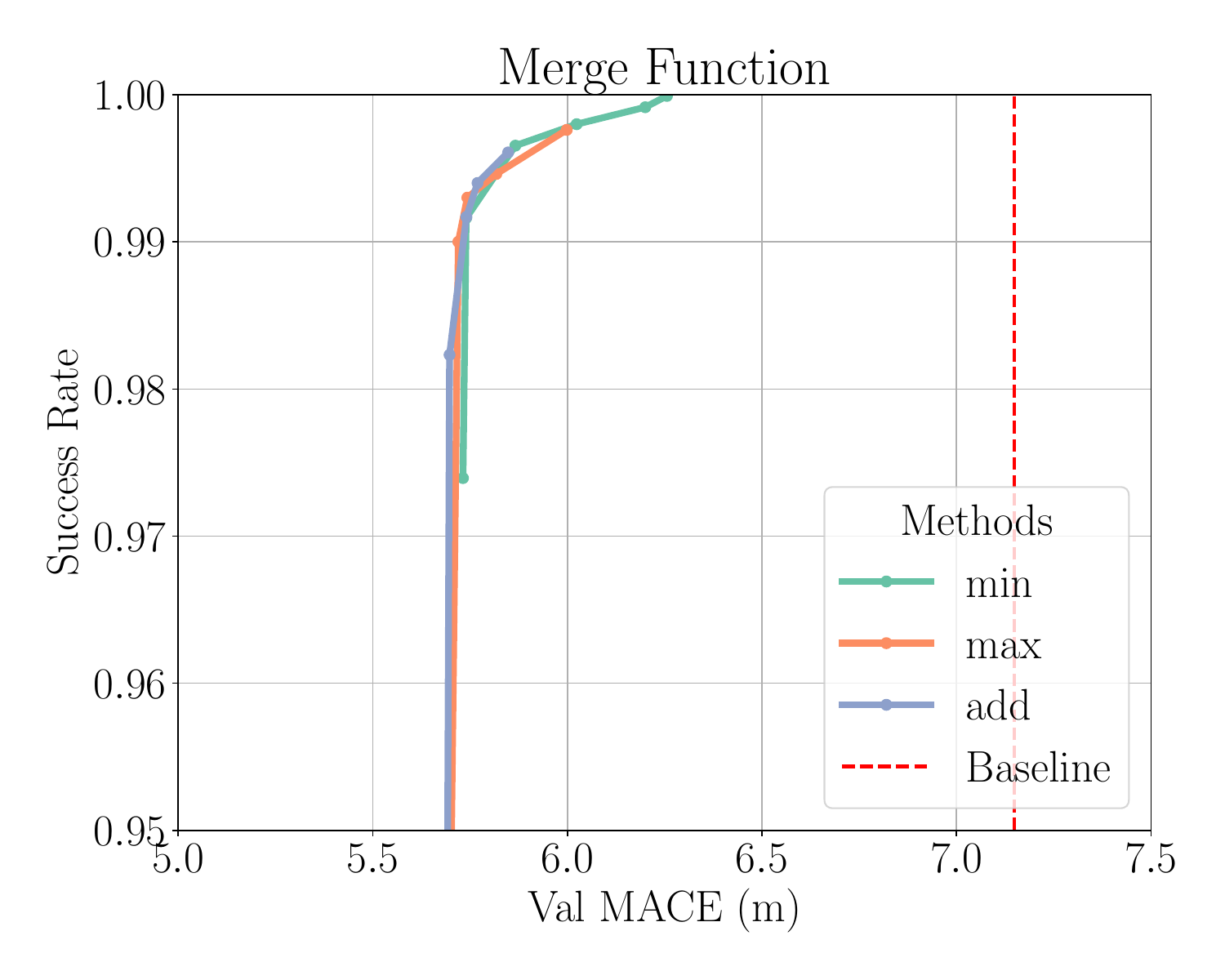}\vspace{-10pt}
    \caption{Merge Function}
    \label{merge}
\end{subfigure}
    \caption{Ablation Study. We use success rate and Validation (Val) MACE metrics to ablate the training and evaluation settings. \textit{Baseline} indicates STHN~\cite{STHN} two-stage baseline performance.}
    \label{ablation}
\vspace{-15pt}
\end{figure*}

\subsection{Deep Ensembles (DE) and Merge Functions} To comprehensively account for both data uncertainty and model uncertainty, we employ Deep Ensembles (DE) for model uncertainty. We train $N_m$ models with different random seeds. During evaluation, we aggregate the predictions and calculate standard deviations as a measure of uncertainty
\begin{equation}
     U^\textrm{DE}_{RS\rightarrow RT} = \textrm{std}(\mathcal{D}^\prime_{RS\rightarrow RT}), 
     \end{equation}
     \begin{equation}\quad\mathcal{D}^\prime_{RS\rightarrow RT} = \{D_{RS\rightarrow RT,m},~m=1,\cdots,N_m\},
\end{equation}
where $U^\textrm{TTA}_{RS\rightarrow RT}\in \mathbb{R}^{2\times4}$ denotes the model uncertainty, and $D_{RS\rightarrow RT,m}$ represents the predicted displacement from the $m^\text{th}$ trained model. Therefore, we obtain data and model uncertainty for each corner and apply a merge function $f(\cdot)$ to evaluate the total uncertainty
\begin{equation}
     U^\textrm{total}_{RS\rightarrow RT} = f(U^\textrm{TTA}_{RS\rightarrow RT}, U^\textrm{DE}_{RS\rightarrow RT}),
\end{equation}
where $U^\textrm{total}_{RS\rightarrow RT}$ represents the total uncertainty, and $f(\cdot)$ is a strategy that utilizes either the minimum/maximum values of data and model uncertainty or their sum. 

To reduce the computational cost of extra samples, we propose an \textit{early stopping} mechanism. It uses all samples for the first $k$ iterations for UE, then switches to a single sample solely for homography estimation, maintaining accuracy while significantly cutting computational overhead.

\section{Experimental Setup}~\label{sec:experiment_setting}
\textbf{Dataset.} We use the Boson-nighttime real-world thermal dataset~\cite{stl, STHN}, which includes paired satellite-thermal images and unpaired satellite images. The 8-bit thermal images were captured with a Boson Thermal Camera during nighttime flights (9:00 PM to 4:00 AM), covering landscapes such as deserts, farms, and roads over $33~\si{km^2}$ for thermal and $216~\si{km^2}$ for satellite images. The dataset uses Bing Satellite Maps, with thermal images ($W_T=512$) aligned to satellite images ($W_S=1536$). It contains $10$K training, $13$K validation, and $27$K test pairs, with test data from different regions and times. Additionally, there are $160$K unpaired satellite images for optional thermal synthesis, excluding validation and test regions to evaluate generalization.

\textbf{Baselines.} We use the following DHE baselines: DHN \cite{detone2016deep}, IHN \cite{cao2022iterative}, and STHN~\cite{STHN}. DHN introduces DHE with four-corner displacement, while IHN adds an iterative approach with a correlation module. STHN uses TGM~\cite{stl} to synthesize thermal images and a two-stage method for DHE in TG. We apply UE only for the first stage. The Direct Modeling (DM) method~\cite{feng2021review} is our data uncertainty baseline.

\textbf{Metrics.} We employ the \textit{Mean Average Corner Error (MACE)} to evaluate homography estimation accuracy and the \textit{Center Error (CE)} to assess geo-localization accuracy. MACE~\cite{detone2016deep, cao2022iterative} is calculated as the average Euclidean distance between predicted and ground truth corners. CE~\cite{STHN} measures the Euclidean distance between predicted and ground truth centers. The \textit{Distance of Centers ($D_C$)} indicates the maximum center distance between thermal and satellite images, with smaller $D_C$ indicating high-frequency localization and larger $D_C$ indicating low-frequency localization. For UE, we use the \textit{Success Rate (SR)}, quantifying the proportion of non-rejected samples.

\textbf{Implementation Details.}\label{imp}
We set the iteration numbers for the homography networks to $K_1=6$ for the first stage and $K_2=6$ for the second stage, with a resizing width $W_R$ of $256$ and a decay factor $\gamma$ of $0.85$. We first train $F_H$ for $100$k steps without CropTTA, then fine-tune it for an additional $200$K steps with CropTTA. For two-stage methods, we apply bounding box augmentation~\cite{STHN}, perturbing the first-stage results by 64 pixels. During the evaluation, the bounding box width $W_B$ is expanded by 64 pixels. All uncertainty estimation methods are evaluated with 5 samples. $F_H$ is trained using the AdamW optimizer~\cite{loshchilov2017decoupled}, with a linear decay scheduler and a maximum learning rate of $1\times10^{-4}$.
\vspace{-5pt}

\begin{table*}[]
    \centering
        \caption{Comparison of test MACE (m), CE (m), and Success Rates (SR) across different UE methods with DHE baselines at $W_S=1536$. All baselines were trained with real and synthesized thermal data. $^\dagger$~denotes methods with a narrow standard deviation range, consistently yielding $100\%$ success rates.}
        \resizebox{\linewidth}{!}{
    \begin{tabular}{lcccccccccccccc}
    \toprule
       \multirow{2}{4em}{DHE Methods} &\multirow{2}{3em}{UE Methods}&\multirow{2}{5em}{Uncertainty} & &\multicolumn{3}{c}{$D_C=128~\si{m}$}& & \multicolumn{3}{c}{$D_C=256~\si{m}$} & & \multicolumn{3}{c}{$D_C=512~\si{m}$}\\
        \cline{5-7}\cline{9-11}\cline{13-15}\vspace{-9pt}\\
        & & & & MACE& CE &SR & & MACE & CE & SR & & MACE & CE &SR\\
         \midrule
         \multirow{5}{3em}{DHN~\cite{detone2016deep}} & -& - & &73.60 &73.58 & 100\% & & 171.93& 171.02 & 100\% & &342.41 & 341.11 & 100\%\\
         & DE & model & &\textbf{61.91} & \textbf{61.90} & 97.3\% & &162.77 &162.75 & 96.4\% & & 346.26 & 346.21 & 93.6\% \\
         & DM & data  & &66.77 &66.76 &99.3\% &  &164.19 &164.17  & 99.7\% & &\textbf{335.81} & \textbf{335.77} & 91.7\% \\
         & CropTTA & data & &64.18 &62.78 & 98.6\% & &\underline{162.04} &\underline{161.82} & 95.1\% & & 337.90 & 337.84 & 91.5\%\\
         & CropTTA + DE & data + model & &\underline{64.09} &\underline{62.70} &96.6\% & &\textbf{161.84}  &  \textbf{161.62}& 96.1\% & & \underline{336.41} &\underline{336.47} & 93.6\%\\
         \midrule
         \multirow{5}{3em}{IHN~\cite{cao2022iterative}} & -& -& &7.27 & 7.24& 100\% & & 16.78 & 16.42 & 100\% & & 16.42& 15.90 & 100\%\\
          & DE & model & &\textbf{6.07} &\textbf{6.06}  & 97.9\% & & 12.31 &12.13 & 97.6\% & &13.73  & 13.37 & 94.5\%\\
          & DM & data  & &\underline{7.02} &\underline{6.99} &100\%$^\dagger$ & &\underline{11.81} &11.40 & 100\%$^\dagger$ & &11.48 & 11.14& 94.1\% \\
         & CropTTA & data & &7.46 &7.47 & 97.4\% & &11.91 &\underline{10.80} & 97.5\%& &\textbf{9.27} & \textbf{8.06}& 95.0\% \\
         & CropTTA + DE & data + model & &7.25 &7.26 & 95.7\% & &\textbf{11.57} &\textbf{10.44} & 97.1\% & &\underline{10.67} & \underline{9.41} & 93.8\%\\
          \midrule
         \multirow{5}{3em}{STHN~\cite{STHN}} & -& - & &\textbf{7.51} &\textbf{6.66} & 100\% & &14.99&14.34 &100\% & &12.70& 12.12 &100\% \\
          & DE & model & &9.45 &8.72 &98.1\% & &9.98 &9.09 & 95.3\%& &8.29 &7.58  & 97.3\%\\
          & DM & data  & &9.75 &8.68 & 100\%$^\dagger$ & &13.64 & 12.91& 100\%$^\dagger$ & &11.35 &10.64  & 100\%$^\dagger$ \\
         & CropTTA & data & &8.26 &7.75 & 98.5\%& &\underline{7.85} &\underline{7.31} & 95.8\% & &\underline{7.93} &\underline{7.25} & 97.5\%\\
         & CropTTA + DE & data + model & &\underline{8.16} &\underline{7.65} &98.1\% & &\textbf{7.50} &\textbf{6.97} & 94.5\% & &\textbf{7.83} & \textbf{7.15} & 97.0\%\\

    \bottomrule
    \end{tabular}}
    \label{baseline}
    \vspace{-10pt}
\end{table*}

\begin{table}
    \centering
    \caption{Comparison of inference time (\si{ms}) for different UE methods with or without early stopping. We evaluate with 5 samples and in an NVIDIA RTX 2080Ti GPU.}
        \resizebox{0.48\textwidth}{!}{
    \begin{tabular}{lccccc}
    \toprule
       Methods & Early Stopping & w/o UE & CropTTA & DE & CropTTA + DE\\
         \midrule
         \multirow{2}{4em}{IHN~\cite{cao2022iterative}} & \xmark & 35.2 & 64.6 & 114.6 & 164.2  \\
         & \checkmark & - & 54.6 &63.1 & 92.1\\\midrule
         \multirow{2}{4em}{STHN~\cite{STHN}} &\xmark & 63.9  &87.0 & 130.2  & 186.0\\
         & \checkmark &  - &78.2 & 81.9 & 118.6\\
    \bottomrule
    \end{tabular}}
    \label{time}
    \vspace{-15pt}
\end{table}

\section{Results}~\label{sec:resutls}

\vspace{-20pt}
\subsection{Ablation Study}\label{sec:ablation}
In this section, we conduct an ablation study (Fig.~\ref{ablation}) to assess the performance impact of our methods' design. We use the STHN two-stage model~\cite{STHN} with $W_S=1536$ and $D_C=512~\si{m}$ as the baseline. We plot the success rate against the Validation (Val) MACE metric with multiple thresholds.

\textbf{Number of Training Crops.} Fig.~\ref{mct} displays the performance of models trained with different numbers of training crops. The results indicate that models trained with $N_C=5$ achieve a higher success rate and lower error compared to those trained with $N_C=1$ (only the original image without crop augmentation) and $N_C=3$.

\textbf{CropTTA Sampling Method.} Fig.~\ref{croptta} compares different sampling methods and crop offsets for CropTTA. To minimize the randomness of random sampling, we average the results of five random seeds for each random sampling method. The results indicate that random sampling generally outperforms grid sampling, with random sampling using $o_c = 32$ yielding the best performance.

\begin{figure}
    \centering
\includegraphics[width=0.35\textwidth,height=0.35\textwidth]{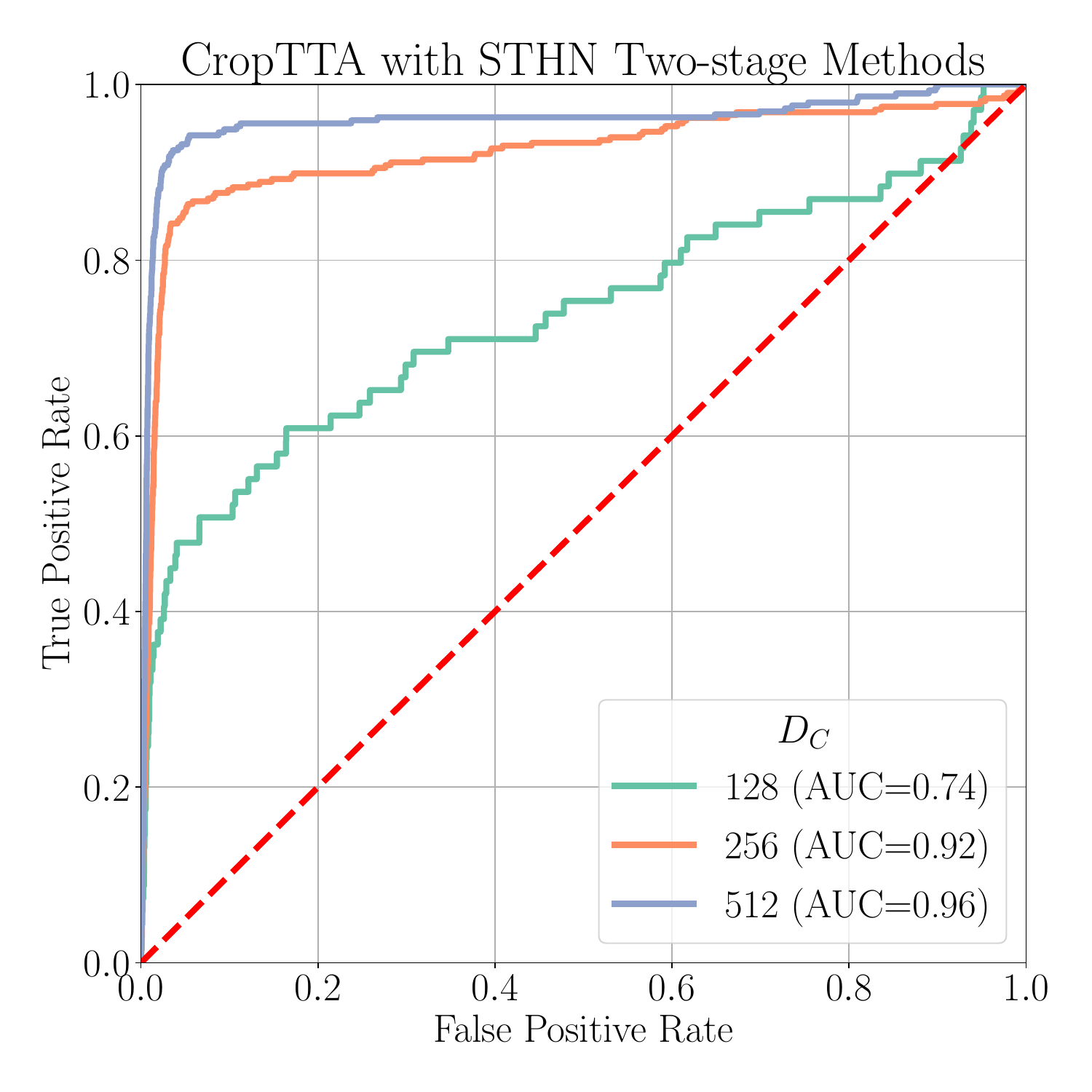}\vspace{-10pt}
   \caption{ROC curves for CropTTA with STHN two-stage methods~\cite{STHN} across different $D_C$, with predictions exceeding $25~\si{m}$ MACE considered as expected rejected predictions.}\label{roc}
   \vspace{-15pt}
\end{figure}

\begin{figure}
    \centering
\includegraphics[width=0.35\textwidth,height=0.35\textwidth]{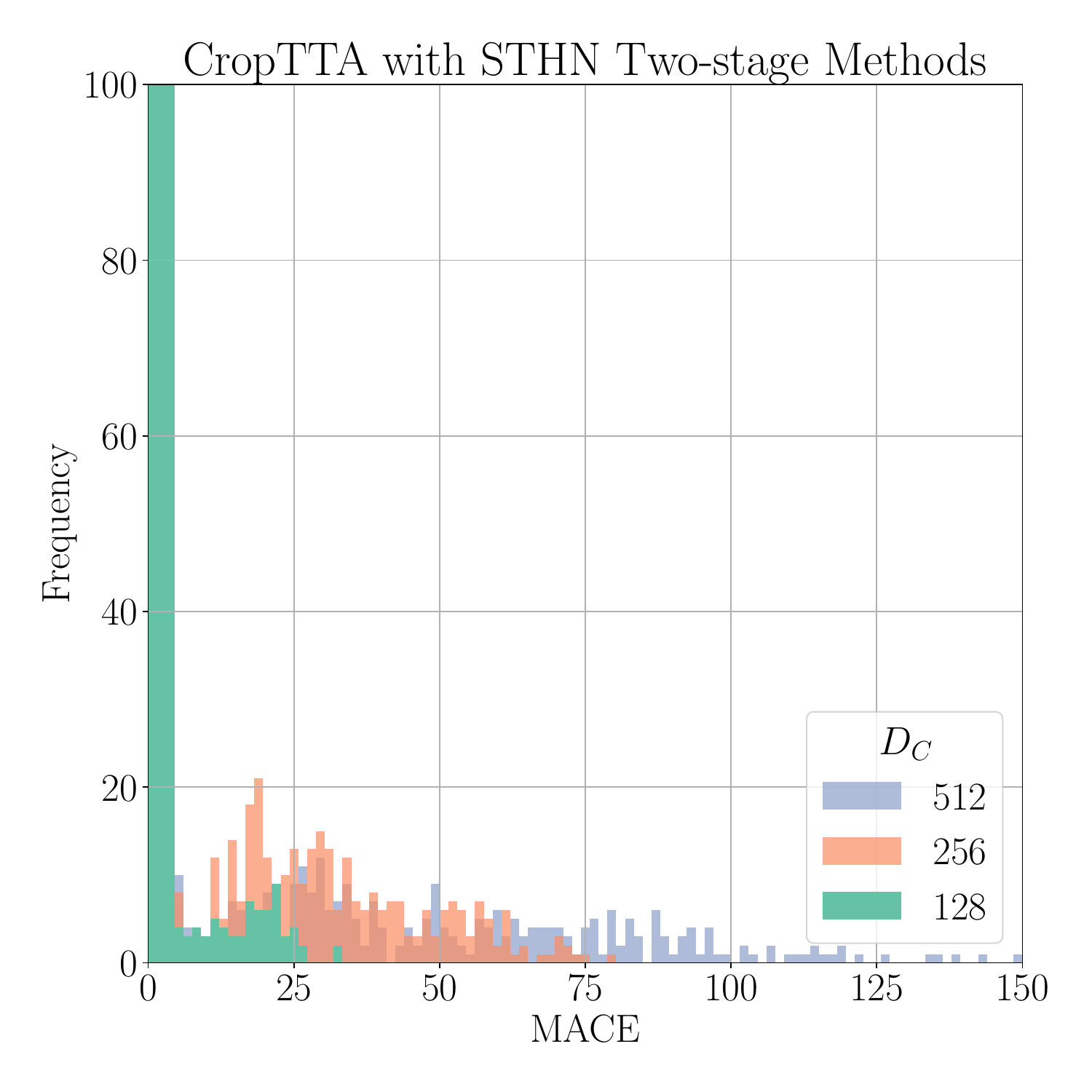}\vspace{-10pt}
   \caption{MACE histogram for CropTTA with STHN two-stage methods across different $D_C$}\label{hist}
   \vspace{-15pt}
\end{figure}

\begin{figure*}[]
\centering
\begin{subfigure}[b]{0.15\textwidth}
    \includegraphics[width=\textwidth]{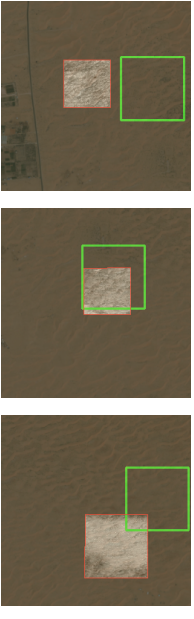}
    \vspace{-20pt}
    \caption{}
        \vspace{-5pt}
\end{subfigure}
\hspace{0.04em}
\begin{subfigure}[b]{0.15\textwidth}
    \includegraphics[width=\textwidth]{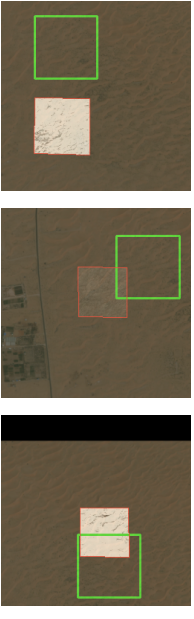}
    \vspace{-20pt}
    \caption{}
        \vspace{-5pt}
\end{subfigure}
\hspace{0.04em}
\begin{subfigure}[b]{0.15\textwidth}
    \includegraphics[width=\textwidth]{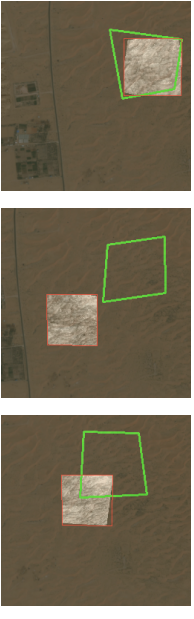}
    \vspace{-20pt}
    \caption{}
        \vspace{-5pt}
\end{subfigure}
\hspace{0.04em}
\begin{subfigure}[b]{0.15\textwidth}
    \includegraphics[width=\textwidth]{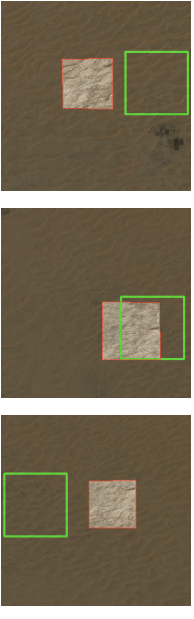}
    \vspace{-20pt}
    \caption{}
    \vspace{-5pt}
\end{subfigure}
\hspace{0.04em}
\begin{subfigure}[b]{0.15\textwidth}
    \includegraphics[width=\textwidth]{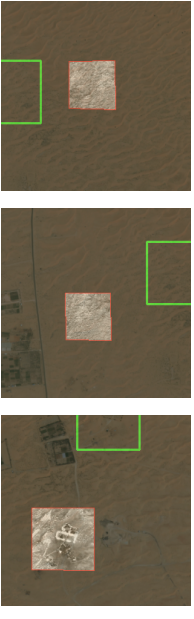}
    \vspace{-20pt}
    \caption{}
        \vspace{-5pt}
\end{subfigure}
\hspace{0.04em}
\begin{subfigure}[b]{0.15\textwidth}
    \includegraphics[width=\textwidth]{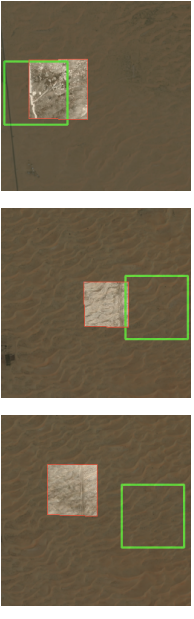}
    \vspace{-20pt}
    \caption{}
    \vspace{-5pt}
\end{subfigure}
\caption{CropTTA detected failure cases with the STHN two-stage method. Thermal images overlap with satellite images, showing \textcolor{green}{ground truth} and \textcolor{red}{predicted} displacements. Thermal images are overlaid on \textcolor{red}{predicted} displacements on the satellite imagery for visualization. Categories from left to right: (a) textureless thermal features, (b) corrupted thermal images, (c) geometric noise, (d) self-similar satellite maps, (e) thermal images exceeding search regions, and (f) outdated satellite maps.}
    \label{vis}
    \vspace{-15pt}
\end{figure*}

\textbf{Aggregation Methods.} Fig.~\ref{agg} compares model performance with different aggregation methods. \textit{Original} uses the original image result, while \textit{mean} averages results from all samples. It shows that DE performs better with the \textit{mean} method, while CropTTA prefers the \textit{original} method, likely due to cropped images containing partial information.

\textbf{Sample Numbers.} Fig.~\ref{sample} illustrates the evaluation results for different sample numbers. The curves indicate that the performance of CropTTA and DE converge when the sample numbers exceed 4 and 3, respectively. This suggests the minimal sample numbers required for optimal performance.

\textbf{Early Stopping.} Fig.~\ref{cs} illustrates the impact of early stopping on UE across different iteration numbers for $F_H$, assuming $F_H$ is an iterative model~\cite{cao2022iterative, STHN}. The results suggest that early stopping can effectively enhance the efficiency of iterative methods without compromising performance.

\textbf{Merge Function.} Fig.~\ref{merge} presents the results of various merge functions used to combine CropTTA and DE. The \textit{max} and \textit{add} methods exhibit similar performance, while the \textit{min} method achieves higher success rates but also higher error. Combining CropTTA and DE allows us to have a comprehensive UE for better performance. We choose \textit{max} as our default merge function.

\subsection{Comparison with Baselines}\label{sec:baseline}
Table~\ref{baseline} evaluates the performance of various UE methods. Our results show that CropTTA enhances alignment accuracy for large $D_C$ values (low-frequency localization). However, for $D_C = 128~\si{m}$ (high-frequency localization), baselines without UE outperform, indicating inherent lower bounds of MACE and CE. Similar errors persist for $D_C = 256~\si{m}$ and $512~\si{m}$ with UE, even after removing high-uncertainty samples. Notably, IHN and STHN achieve superior performance for $D_C = 256~\si{m}$ and $512~\si{m}$, maintaining success rates above 95\%. Figure~\ref{roc} shows the ROC curves for STHN with CropTTA, where $D_C = 256\si{m}$ and $512~\si{m}$ yield higher true positive rates than $D_C = 128~\si{m}$. This is supported by the long-tail error distribution in the MACE histogram (Fig.~\ref{hist}), showing improved UE performance.

Comparing UE methods, CropTTA generally outperforms or matches DE and DM, except when DHN has high errors or IHN under $D_C=128~\si{m}$. DM struggles to detect data uncertainty in low-error results, leading to uncertainty underestimation, while CropTTA effectively handles these cases. Combining CropTTA and DE improves performance when both estimate well but suffers if either has high errors.

\textbf{Inference Time.} Table~\ref{time} compares inference times of various UE methods, showing CropTTA is more efficient than DE in both one-stage and two-stage methods. Early stopping also reduces the inference time with more samples, making it practical for real-time TG applications.

\subsection{Failure Detection}\label{sec:robustness}
Fig.~\ref{vis} provides a qualitative analysis of failures detected by CropTTA, highlighting six categories of high data uncertainty samples. Textureless thermal images show low contrast or flat features without landmarks. Corrupted thermal images had brightness adjusted for overexposure and underexposure, while geometric noise was added by shifting corners by up to 64 pixels. Self-similar satellite maps, like desert dunes, exhibit repetitive patterns. Thermal images extending beyond the search region had displacements partially outside satellite boundaries. The outdated satellite map in 2020 is compared with thermal images captured in 2021, revealing new roads and farms. CropTTA effectively detects failures due to textureless, corrupted, and out-of-range images, self-similar patterns, outdated maps, and geometric noise.

\section{Conclusions}~\label{sec:conclusions}
In this study, we introduced \textit{UASTHN}, a novel uncertainty estimation method for Deep Homography Estimation (DHE) in thermal geo-localization, which significantly enhances the reliability of UAV outdoor nighttime localization and navigation. Our CropTTA strategy has demonstrated robust failure detection in challenging TG scenarios. 

Future works will explore the development of an adaptive rejection mechanism and evaluate it on diverse datasets, including daytime, seasonal variations, and dynamic objects.




\bibliographystyle{IEEEtran} 
\bibliography{mybib}

\begin{thebibliography}{10}
\providecommand{\url}[1]{#1}
\csname url@samestyle\endcsname
\providecommand{\newblock}{\relax}
\providecommand{\bibinfo}[2]{#2}
\providecommand{\BIBentrySTDinterwordspacing}{\spaceskip=0pt\relax}
\providecommand{\BIBentryALTinterwordstretchfactor}{4}
\providecommand{\BIBentryALTinterwordspacing}{\spaceskip=\fontdimen2\font plus
\BIBentryALTinterwordstretchfactor\fontdimen3\font minus \fontdimen4\font\relax}
\providecommand{\BIBforeignlanguage}[2]{{%
\expandafter\ifx\csname l@#1\endcsname\relax
\typeout{** WARNING: IEEEtran.bst: No hyphenation pattern has been}%
\typeout{** loaded for the language `#1'. Using the pattern for}%
\typeout{** the default language instead.}%
\else
\language=\csname l@#1\endcsname
\fi
#2}}
\providecommand{\BIBdecl}{\relax}
\BIBdecl

\bibitem{info10110349}
D.~C. Tsouros, S.~Bibi, and P.~G. Sarigiannidis, ``A review on uav-based applications for precision agriculture,'' \emph{Information}, vol.~10, no.~11, 2019.

\bibitem{drones6110347}
L.~Morando, C.~T. Recchiuto, J.~Calla, P.~Scuteri, and A.~Sgorbissa, ``Thermal and visual tracking of photovoltaic plants for autonomous uav inspection,'' \emph{Drones}, vol.~6, no.~11, 2022.

\bibitem{atif2021uav}
M.~Atif, R.~Ahmad, W.~Ahmad, L.~Zhao, and J.~J. Rodrigues, ``Uav-assisted wireless localization for search and rescue,'' \emph{IEEE Systems Journal}, vol.~15, no.~3, pp. 3261--3272, 2021.

\bibitem{rao2022quadformer}
P.~P. Rao, F.~Qiao, W.~Zhang, Y.~Xu, Y.~Deng, G.~Wu, and Q.~Zhang, ``Quadformer: Quadruple transformer for unsupervised domain adaptation in power line segmentation of aerial images,'' \emph{arXiv preprint arXiv:2211.16988}, 2022.

\bibitem{powerline}
J.~Xing, G.~Cioffi, J.~Hidalgo-Carrió, and D.~Scaramuzza, ``Autonomous power line inspection with drones via perception-aware mpc,'' in \emph{IEEE/RSJ International Conference on Intelligent Robots and Systems (IROS)}, 2023, pp. 1086--1093.

\bibitem{saviolo2023unifying}
A.~Saviolo, P.~Rao, V.~Radhakrishnan, J.~Xiao, and G.~Loianno, ``Unifying foundation models with quadrotor control for visual tracking beyond object categories,'' in \emph{IEEE International Conference on Robotics and Automation (ICRA)}, 2024, pp. 7389--7396.

\bibitem{review_avl}
A.~Couturier and M.~A. Akhloufi, ``A review on absolute visual localization for uav,'' \emph{Robotics and Autonomous Systems}, vol. 135, p. 103666, 2021.

\bibitem{foundloc}
Y.~He, I.~Cisneros, N.~Keetha, J.~Patrikar, Z.~Ye, I.~Higgins, Y.~Hu, P.~Kapoor, and S.~Scherer, ``Foundloc: Vision-based onboard aerial localization in the wild,'' \emph{arXiv preprint arXiv:2310.16299}, 2023.

\bibitem{vgscience}
A.~T. Fragoso, C.~T. Lee, A.~S. McCoy, and S.-J. Chung, ``A seasonally invariant deep transform for visual terrain-relative navigation,'' \emph{Science Robotics}, vol.~6, no.~55, p. eabf3320, 2021.

\bibitem{directalign3}
B.~Patel, T.~D. Barfoot, and A.~P. Schoellig, ``Visual localization with google earth images for robust global pose estimation of uavs,'' in \emph{IEEE International Conference on Robotics and Automation (ICRA)}, 2020, pp. 6491--6497.

\bibitem{imgregistration}
M.~Shan, F.~Wang, F.~Lin, Z.~Gao, Y.~Z. Tang, and B.~M. Chen, ``Google map aided visual navigation for uavs in gps-denied environment,'' in \emph{IEEE International Conference on Robotics and Biomimetics (ROBIO)}, 2015, pp. 114--119.

\bibitem{lee2024caltech}
C.~Lee, M.~Anderson, N.~Raganathan, X.~Zuo, K.~Do, G.~Gkioxari, and S.-J. Chung, ``Caltech aerial rgb-thermal dataset in the wild,'' \emph{arXiv preprint arXiv:2403.08997}, 2024.

\bibitem{stl}
J.~Xiao, D.~Tortei, E.~Roura, and G.~Loianno, ``Long-range uav thermal geo-localization with satellite imagery,'' in \emph{IEEE/RSJ International Conference on Intelligent Robots and Systems (IROS)}, 2023, pp. 5820--5827.

\bibitem{STHN}
J.~Xiao, N.~Zhang, D.~Tortei, and G.~Loianno, ``Sthn: Deep homography estimation for uav thermal geo-localization with satellite imagery,'' \emph{IEEE Robotics and Automation Letters}, vol.~9, no.~10, pp. 8754--8761, 2024.

\bibitem{lecun2015deep}
Y.~LeCun, Y.~Bengio, and G.~Hinton, ``Deep learning,'' \emph{nature}, vol. 521, no. 7553, pp. 436--444, 2015.

\bibitem{detone2016deep}
D.~DeTone, T.~Malisiewicz, and A.~Rabinovich, ``Deep image homography estimation,'' \emph{arXiv preprint arXiv:1606.03798}, 2016.

\bibitem{cao2022iterative}
S.-Y. Cao, J.~Hu, Z.~Sheng, and H.-L. Shen, ``Iterative deep homography estimation,'' in \emph{Proceedings of the IEEE/CVF Conference on Computer Vision and Pattern Recognition (CVPR)}, 2022, pp. 1879--1888.

\bibitem{nguyen2018unsupervised}
T.~Nguyen, S.~W. Chen, S.~S. Shivakumar, C.~J. Taylor, and V.~Kumar, ``Unsupervised deep homography: A fast and robust homography estimation model,'' \emph{IEEE Robotics and Automation Letters}, vol.~3, no.~3, pp. 2346--2353, 2018.

\bibitem{shao2021localtrans}
R.~Shao, G.~Wu, Y.~Zhou, Y.~Fu, L.~Fang, and Y.~Liu, ``Localtrans: A multiscale local transformer network for cross-resolution homography estimation,'' in \emph{Proceedings of the IEEE/CVF international conference on computer vision (CVPR)}, 2021, pp. 14\,890--14\,899.

\bibitem{pmlr-v155-achermann21a}
F.~Achermann, A.~Kolobov, D.~Dey, T.~Hinzmann, J.~J. Chung, R.~Siegwart, and N.~Lawrance, ``Multipoint: Cross-spectral registration of thermal and optical aerial imagery,'' in \emph{Proceedings of the 2020 Conference on Robot Learning}, ser. Proceedings of Machine Learning Research, J.~Kober, F.~Ramos, and C.~Tomlin, Eds., vol. 155.\hskip 1em plus 0.5em minus 0.4em\relax PMLR, 16--18 Nov 2021, pp. 1746--1760.

\bibitem{electronics12040788}
Y.~Luo, X.~Wang, Y.~Wu, and C.~Shu, ``Infrared and visible image homography estimation using multiscale generative adversarial network,'' \emph{Electronics}, vol.~12, no.~4, 2023.

\bibitem{electronics12214441}
X.~Wang, Y.~Luo, Q.~Fu, Y.~He, C.~Shu, Y.~Wu, and Y.~Liao, ``Coarse-to-fine homography estimation for infrared and visible images,'' \emph{Electronics}, vol.~12, no.~21, 2023.

\bibitem{gawlikowski2023survey}
J.~Gawlikowski, C.~R.~N. Tassi, M.~Ali, J.~Lee, M.~Humt, J.~Feng, A.~Kruspe, R.~Triebel, P.~Jung, R.~Roscher \emph{et~al.}, ``A survey of uncertainty in deep neural networks,'' \emph{Artificial Intelligence Review}, vol.~56, no. Suppl 1, pp. 1513--1589, 2023.

\bibitem{kendall2017uncertainties}
A.~Kendall and Y.~Gal, ``What uncertainties do we need in bayesian deep learning for computer vision?'' in \emph{Advances in Neural Information Processing Systems}, vol.~30, 2017.

\bibitem{ABDAR2021243}
M.~Abdar, F.~Pourpanah, S.~Hussain, D.~Rezazadegan, L.~Liu, M.~Ghavamzadeh, P.~Fieguth, X.~Cao, A.~Khosravi, U.~R. Acharya, V.~Makarenkov, and S.~Nahavandi, ``A review of uncertainty quantification in deep learning: Techniques, applications and challenges,'' \emph{Information Fusion}, vol.~76, pp. 243--297, 2021.

\bibitem{shanmugam2021better}
D.~Shanmugam, D.~Blalock, G.~Balakrishnan, and J.~Guttag, ``Better aggregation in test-time augmentation,'' in \emph{Proceedings of the IEEE/CVF international conference on computer vision}, 2021, pp. 1214--1223.

\bibitem{kimura2021understanding}
M.~Kimura, ``Understanding test-time augmentation,'' in \emph{International Conference on Neural Information Processing}.\hskip 1em plus 0.5em minus 0.4em\relax Springer, 2021, pp. 558--569.

\bibitem{kim2020learning}
I.~Kim, Y.~Kim, and S.~Kim, ``Learning loss for test-time augmentation,'' in \emph{Advances in Neural Information Processing Systems}, vol.~33, 2020, pp. 4163--4174.

\bibitem{zhang2022memo}
M.~Zhang, S.~Levine, and C.~Finn, ``Memo: Test time robustness via adaptation and augmentation,'' in \emph{Advances in Neural Information Processing Systems}, vol.~35, 2022, pp. 38\,629--38\,642.

\bibitem{lakshminarayanan2017simple}
B.~Lakshminarayanan, A.~Pritzel, and C.~Blundell, ``Simple and scalable predictive uncertainty estimation using deep ensembles,'' in \emph{Advances in neural information processing systems}, vol.~30, 2017.

\bibitem{NEURIPS2021_a70dc404}
R.~Rahaman and A.~Thiery, ``Uncertainty quantification and deep ensembles,'' in \emph{Advances in Neural Information Processing Systems}, M.~Ranzato, A.~Beygelzimer, Y.~Dauphin, P.~Liang, and J.~W. Vaughan, Eds., vol.~34.\hskip 1em plus 0.5em minus 0.4em\relax Curran Associates, Inc., 2021, pp. 20\,063--20\,075.

\bibitem{fort2019deep}
S.~Fort, H.~Hu, and B.~Lakshminarayanan, ``Deep ensembles: A loss landscape perspective,'' \emph{arXiv preprint arXiv:1912.02757}, 2019.

\bibitem{abe2022deep}
T.~Abe, E.~K. Buchanan, G.~Pleiss, R.~Zemel, and J.~P. Cunningham, ``Deep ensembles work, but are they necessary?'' in \emph{Advances in Neural Information Processing Systems}, vol.~35, 2022, pp. 33\,646--33\,660.

\bibitem{zhang2022hvc}
H.~Zhang and Y.~Ling, ``Hvc-net: Unifying homography, visibility, and confidence learning for planar object tracking,'' in \emph{European Conference on Computer Vision}.\hskip 1em plus 0.5em minus 0.4em\relax Springer, 2022, pp. 701--718.

\bibitem{xu2022cuahn}
Y.~Xu and G.~C. de~Croon, ``Cuahn-vio: Content-and-uncertainty-aware homography network for visual-inertial odometry,'' \emph{arXiv preprint arXiv:2208.13935}, 2022.

\bibitem{ABDELAZIZ2015103}
Y.~Abdel-Aziz, H.~Karara, and M.~Hauck, ``Direct linear transformation from comparator coordinates into object space coordinates in close-range photogrammetry,'' \emph{Photogrammetric Engineering \& Remote Sensing}, vol.~81, no.~2, pp. 103--107, 2015.

\bibitem{feng2021review}
D.~Feng, A.~Harakeh, S.~L. Waslander, and K.~Dietmayer, ``A review and comparative study on probabilistic object detection in autonomous driving,'' \emph{IEEE Transactions on Intelligent Transportation Systems}, vol.~23, no.~8, pp. 9961--9980, 2021.

\bibitem{loshchilov2017decoupled}
I.~Loshchilov and F.~Hutter, ``Decoupled weight decay regularization,'' in \emph{International Conference on Learning Representations}, 2019.

\end{thebibliography}

\end{document}